\documentclass[12pt, a4paper]{article}
\usepackage{abstract}
\usepackage{fancyhdr}
\usepackage{authblk}
\usepackage{graphicx}
\usepackage[square,sort,comma,numbers]{natbib}
\usepackage{amsfonts,amsmath,amssymb}
\usepackage{bm}
\usepackage{mathtools}

\DeclareMathOperator*{\argmax}{arg\,max}

\DeclarePairedDelimiter{\abs}{\vert}{\vert}

\title{Interpretable Control by Reinforcement Learning}

\author[$\star$]{Daniel Hein}
\author[$\star$]{Steffen Limmer}
\author[$\star$]{Thomas A. Runkler}
\affil[$\star$]{\small Siemens AG, Corporate Technology, Otto-Hahn-Ring 6, 81739 Munich, Germany (e-mail: hein.daniel@siemens.com).\normalsize}

\newcommand{\abstractText}{\noindent
	In this paper, three recently introduced reinforcement learning (RL) methods are used to generate human-interpretable policies for the cart-pole balancing benchmark.
	The novel RL methods learn human-interpretable policies in the form of compact fuzzy controllers and simple algebraic equations.
	The representations as well as the achieved control performances are compared with two classical controller design methods and three non-interpretable RL methods.
	All eight methods utilize the same previously generated data batch and produce their controller offline - without interaction with the real benchmark dynamics.
	The experiments show that the novel RL methods are able to automatically generate well-performing policies which are at the same time human-interpretable.
	Furthermore, one of the methods is applied to automatically learn an equation-based policy for a hardware cart-pole demonstrator by using only human-player-generated batch data.
	The solution generated in the first attempt already represents a successful balancing policy, which demonstrates the methods applicability to real-world problems.
}

\usepackage{hyperref}
\hypersetup{colorlinks=true, urlcolor=blue, linkcolor=blue, citecolor=blue}

\begin{document}
	\bibliographystyle{plainnat}

	
		\maketitle
		\thispagestyle{fancy}
		\lhead{\copyright 2020 the
			authors. This work has been accepted to IFAC for publication under a Creative Commons
			Licence CC-BY-NC-ND.}
		\begin{abstract}
			\abstractText
			\newline
			\newline
		\end{abstract}
	
	
	\section{Introduction}
\label{section:introduction}

The search for interpretable reinforcement learning (RL) policies is of high academic and industrial interest.
Interpretability is defined as the ability to explain or to present in understandable terms to a human.
In the context of machine learning (ML), \citeauthor{doshi:17} argue that the need for interpretability stems from an incompleteness in the problem formalization, creating a fundamental barrier to optimization and evaluation~\citep{doshi:17}.
Since complex real-world tasks in industry are almost never completely testable, enumerating all possible outputs given all possible inputs is infeasible.
Hence, we usually are unable to flag all undesirable outputs.
Especially for industrial systems, domain experts are more likely to deploy autonomously learned controllers if they are understandable and convenient to assess.
Moreover, legal frameworks such as the European Union's General Data Protection Regulation~\citep{eu:16} enforce interpretability of data processing systems.

In the context of this work, interpretability is evaluated via a quantifiable proxy. 
At first, a model class, e.g., fuzzy rules, algebraic equations, or high-level task descriptions, is claimed of being interpretable, and subsequently algorithms are developed in order to optimize within this class.

Since controllers from classical control theory, like pro-portional-integral-derivative (PID) controllers, traditionally have been parameterized by iterating response observation and adapting the respective parameters, they are generally designed to be interpretable for humans.
For this reason, PID control is still a very important and common strategy in order to realize stable, efficient, and understandable control strategies in industry applications.
In 2002, \citeauthor{desborough:02} conducted a survey of more than $11\,000$ controllers in the refining, chemicals, and pulp and paper industries which revealed that 97\% of regulatory controllers had a PID structure~\citep{desborough:02}.

\begin{figure}
	\begin{center}
		\includegraphics[width=3.4in]{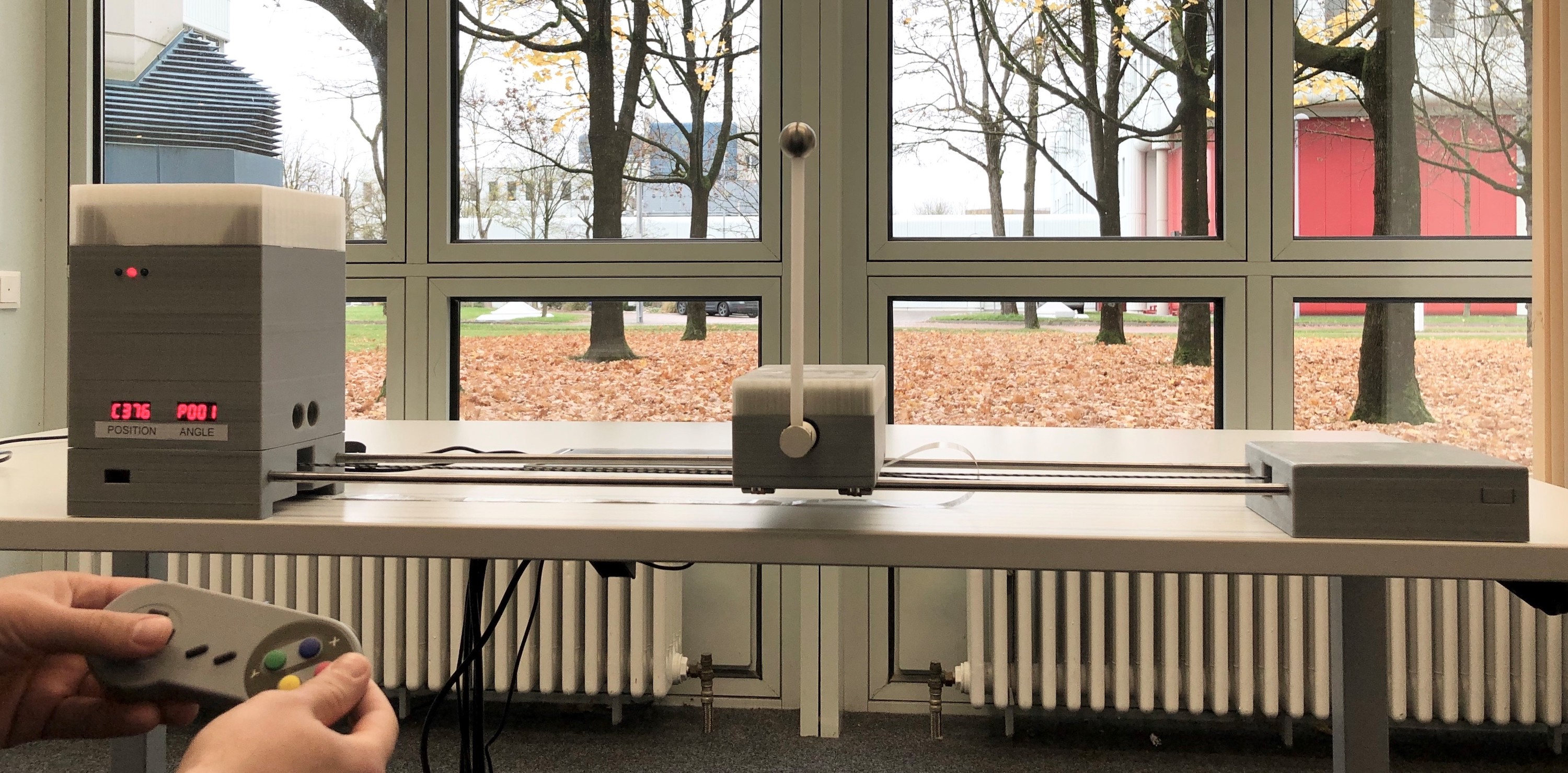}
		\caption{Hardware cart-pole system}
		\label{figure:cebit_pendel}
	\end{center}
\end{figure} 

However, the field of application of regulating PID controllers is rather limited, since they are only linear, symmetric around the setpoint, do not incorporate planning, and cannot utilize direct knowledge of the process, for example.

RL on the other hand, is theoretically capable of solving any Markov decision process (MDP) which makes it applicable to a dramatically broader field of application compared to classical PID control.
Moreover, since RL is not limited to linear control, can incorporate the full observation of the system, and is able to solve planning tasks, it is capable of outperforming PID controllers even in classical industrial applications~\citep{bischoff:13}.

In this paper, we compare two methods from control theory, with three non-interpretable RL methods, and three recently introduced interpretable RL methods.
The eight methods are tested using the same data generated by the cart-pole balancing (CPB) benchmark.
Since for many real-world problems, the cost of interacting with the real environment on a trial-and-error basis is prohibitive, the methods were not allowed to perform exploration on the real benchmark dynamics but had to utilize existing state transition batch data generated by an arbitrary default exploration policy.
To demonstrate the applicability on real-world applications, we learned a controller on batch data generated on a physical cart-pole system (Fig.~\ref{figure:cebit_pendel}).
	
	\section{Model-based Batch Reinforcement Learning}
\label{section:rl}

Generally, RL is distinguished from other computational approaches by its emphasis on an agent learning from direct interaction with its environment.
This approach, which does not rely on exemplary supervision or complete models of the environment, is referred to as  \textit{online learning}.
However, for many real-world problems online learning is prohibited for safety reasons.
For example, it is not advisable to deploy an online RL agent, who starts by applying an arbitrary initial policy, which is subsequently improved by exploitation and exploration, on safety-critical systems like power plants or vehicles.
For this reason, \textit{offline learning} is a more suitable approach for applying RL methods on already collected training data and, if proving useful, existing models, to yield RL policies.

Offline RL is often referred to as \textit{batch learning} because it is based solely on a previously generated batch of transition samples from the environment.
The batch data set contains transition tuples of the form $(\bm{s}_t,\bm{a}_t,\bm{s}_{t+1},r_{t+1})$, where the application of action $\bm{a}_t$ in state $\bm{s}_t$ resulted in a transition to state $\bm{s}_{t+1}$ and yielded a reward $r_{t+1}$.
By applying a batch-mode RL algorithm called \textit{fitted Q iteration} (FQI)~\citep{ernst:05}, an action-value function $q_{\pi}$ can be estimated from the available data batch.
FQI is an algorithm which uses the fixed point property of the Bellman equation~\citep{bellman:62} to perform an  \textit{iterative policy evaluation}.
Let's denote the initial action-value function of policy $\pi$ with $q_0$ and choose its initial approximation arbitrarily.
We can now compute each successive approximation by using the Bellman equation as an update rule:
\begin{equation}
q_{k+1}(\bm{s}_t,\bm{a}_t)\leftarrow r_{t+1}+\gamma q_k(\bm{s}_{t+1},\pi(\bm{s}_{t+1})),
\label{eq:action_value_function_update_step}
\end{equation}
for all transition tuples in the data batch.
Depending on the function approximation applied for $q$, this algorithm is expected to converge to $q_\pi$ as $k\rightarrow\infty$.
The resulting action-value function is used as follows:
\begin{align}
v_\pi(\bm{s})&=q_\pi(\bm{s},\pi(\bm{s})),
\label{eq:model_free_evaluation}
\end{align}
for all states $\bm{s}$ of the state space.

Another way of evaluating a policy offline from a batch of transition samples is referred to as \textit{model-based value estimation}.
In the first step, supervised ML is applied to learn approximate models of the underlying environment from transition tuples $(\bm{s}_t,\bm{a}_t,\bm{s}_{t+1},r_{t+1})$ as follows:
\begin{equation}
\tilde{g}(\bm{s}_t,\bm{a}_t)\leftarrow\bm{s}_{t+1},\quad \tilde{r}(\bm{s}_t,\bm{a}_t,\bm{s}_{t+1})\leftarrow r_{t+1}.
\label{eq:model}
\end{equation}
Using models $\tilde{g}$ and $\tilde{r}$, the value for policy $\pi$ of each state $\bm{s}$ in the data batch can be estimated by computing $\tilde{v}_\pi(\bm{s})$.
Hence, using model-based value estimation means performing trajectory rollouts on system models for different starting states:
\begin{align}  
\tilde{v}_{\pi}(\bm{s}_t) &= \sum_{k=0}^{\infty}\gamma^k \tilde{r}(\bm{s}_{t+k},\bm{a}_{t+k},\bm{s}_{t+k+1}),\label{eq:model_based_evaluation} \\
\textnormal{with}\quad \bm{s}_{t+k+1} & = \tilde{g}(\bm{s}_{t+k},\bm{a}_{t+k}),\quad \bm{a}_{t+k}=\pi(\bm{s}_{t+k}).
\end{align}

\subsection{Population-based reinforcement learning}

Since we are searching for interpretable solutions, the resulting policies have to be represented in an explicit form.
\textit{Policy search} yields explicit policies which can be enforced to be of an interpretable form.
Furthermore, policy search is inherently well-suited for being used with population-based optimization techniques, like particle swarm optimization (PSO) and genetic programming (GP). 

The goal of using policy search for learning interpretable RL policies is to find the best policy among a set of policies that is spanned by a parameter vector $\bm{x}\in \mathcal X$. 
Herein, a policy corresponding to a particular parameter value $\bm{x}$ is denoted by $\pi[\bm{x}]$.
The policy's performance, when starting from $\bm{s}_t$ is measured by the value functions $v_{\pi[\bm{x}]}(\bm{s}_t)$ (Eq.~\eqref{eq:model_free_evaluation}) or $\tilde{v}_{\pi[\bm{x}]}(\bm{s}_t)$ (Eq.~\eqref{eq:model_based_evaluation}).
Furthermore, including only a finite number of $T>1$ future rewards for the model-based value estimation from Eq.~\eqref{eq:model_based_evaluation} yields
\begin{equation}
\begin{aligned}    
\tilde{v}_{\pi[\bm{x}]}(\bm{s}_t) &= \sum_{k=0}^{T-1}\gamma^k \tilde{r}(\bm{s}_{t+k},\bm{a}_{t+k},\bm{s}_{t+k+1}), \\
\textnormal{with}\quad \bm{s}_{t+k+1} & = \tilde{g}(\bm{s}_{t+k},\bm{a}_{t+k}) \quad \textnormal{and}\quad \bm{a}_{t+k}=\pi[\bm{x}](\bm{s}_{t+k}).
\end{aligned}
\label{eq:model_based_value}
\end{equation}

To rate the performance of policy ${\pi[\bm{x}]}$, the value function is used as follows:
\begin{equation}
	\overline{\mathcal{R}}_{\pi[\bm{x}]}=\frac{1}{\abs{\mathcal{S}}}\sum_{\bm{s}\in\mathcal{S}}\tilde{v}_{\pi[\bm{x}]}(\bm{s}),
\label{eq:return}
\end{equation}
with $\overline{\mathcal{R}}_{\pi[\bm{x}]}$ being the average discounted return of ${\pi[\bm{x}]}$ on a representative set of test states $\mathcal{S}$.

\begin{figure}
	\begin{center}
	\includegraphics[width=3.4in]{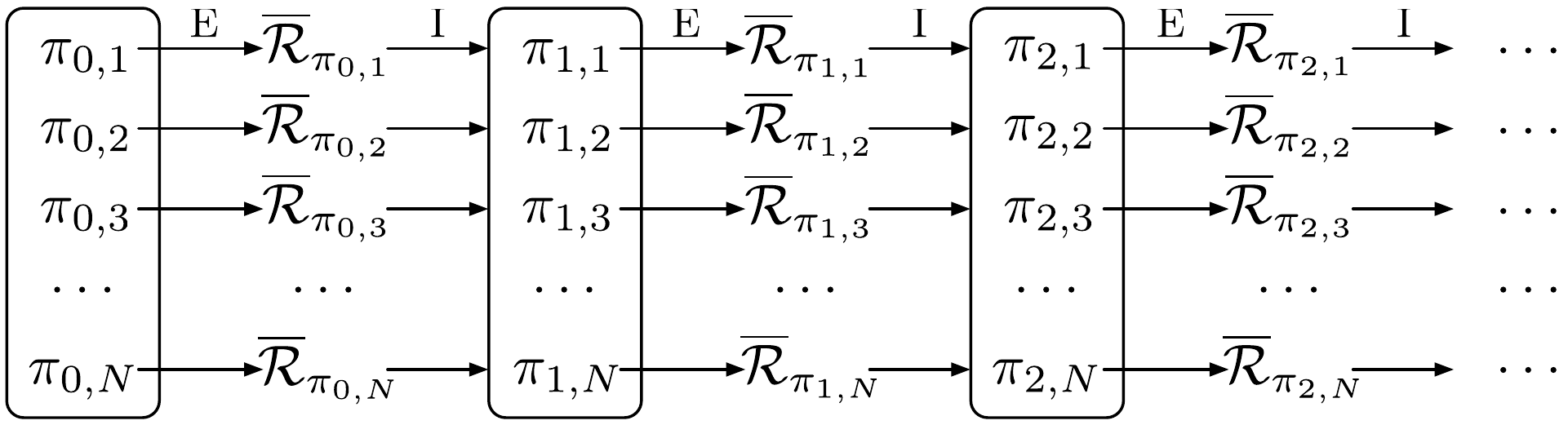}
	\caption{Population-based policy search iterating policy evaluation (E) and policy improvement (I)}
	\label{fig:policy_search_population}
\end{center}
\end{figure}

In population-based RL, the interest lies in populations of policies, which means that multiple different policies exist at the same time in one policy iteration step (Fig.~\ref{fig:policy_search_population}).
Applying population-based evolutionary methods to policy search has already been successfully achieved in the past~\citep{chin:98,gomez:06,chang:07,wilson:18}.
	
	\section{Compared methods}
	\label{section:methods}
	
	\subsection{Linear quadratic regulator (LQR)}
\label{section:lqr}

A well-established control strategy is the linear quadratic regulator (LQR) based on the assumption of a linear dynamical system
\begin{align}\label{equ:lqr_state_trans}
\bm{s}_{t+1} = g(\bm{s}_t, \bm{a}_t) = \bm{U}\bm{s}_{t} + \bm{V}\bm{a}_{t},
\end{align}
where the matrices $\bm{U}$ and $\bm{V}$ are of appropriate dimension and determine the state transition \citep{bertsekas:95}. 
In practical settings, these matrices may either be determined by the governing equations of a physical system or have to be estimated based on observational data. 
In the latter case, the transition matrices are typically estimated from a sequence of state-action pairs of length $T+1$ by defining the matrices $\bm{S} = [\bm{s}_0, \bm{s}_1, \hdots, \bm{s}_T]$, $\bm{A} = [\bm{a}_0, \bm{a}_1, \hdots, \bm{a}_T]$ and $\bm{Y} = [\bm{s}_1, \bm{s}_2, \hdots, \bm{s}_{T+1}]$ and solving the least-squares problem in matrix form given by
\begin{align}
\text{min}_{U,V} \ \lVert \bm{Y} - \bm{S}^T [\bm{U}, \bm{V}] \rVert_F^2.
\end{align}

The LQR approach assumes a \emph{quadratic} cost function of the form
\begin{align}
J(\bm{s}, \bm{a}) = \sum\nolimits_{t=0}^\infty \bm{s}_t^T \bm{Q} \bm{s}_t + \bm{a}_t^T\bm{R} \bm{a}_t,
\end{align}
where the matrices $\bm{Q}$ and $\bm{R}$ are positive semidefinite and account for costs related to the state $\bm{s}_t$ and action $\bm{a}_t$ at time $t$, respectively. 
The restriction to a quadratic cost function leads to favorable analytical properties and can be justified by choosing the matrices $\bm{Q}$, $\bm{R}$ such that they reflect the desire to stabilize the state $\bm{s}$ and action $\bm{a}$ around the origin. 

Under some mild assumptions on the matrices $\bm{U}$,$\bm{V}$,$\bm{Q}$,$\bm{R}$, the optimal controller is \emph{linear} so that the optimal policy can be written as
\begin{align}
\bm{a}_{t} = \pi(\bm{s}_t) = -\bm{K}\bm{s}_t.
\end{align}
Here, the matrix $\bm{K}$ satisfies the \emph{discrete-time algebraic Riccati equation} (DARE) 
\begin{align}\label{equ:lqr_riccati}
\bm{K} = \bm{U}^T ( \bm{K} - \bm{K} \bm{V}( \bm{V}^T \bm{K} \bm{V} + \bm{R})^{-1} \bm{V}^T \bm{K}) \bm{U} + \bm{Q}
\end{align}
and can be found by efficient solvers that are built around optimized linear algebra subroutines.
 
It is important to highlight the assumptions and idealization of the LQR approach compared to many practical control problems.
Firstly, the approach is based on an idealized \emph{linear} state transition function, cf. Eq.~ \eqref{equ:lqr_state_trans}, which is often violated due to non-linear physical effects such as friction and force ripple. 
Secondly, the applied action $\bm{a}$ is penalized but assumed to be \emph{unconstrained} so that in principle arbitrary forces may be applied in order to stabilize the system.

	\subsection{Proportional-integral-derivative (PID) controller}
\label{section:pid}

A PID controller is a control loop feedback mechanism which continuously calculates an error value as the difference between a setpoint and a measured process variable.
Based on proportional, integral, and derivative terms of this error value, a corrective action is computed and applied to the system.
Each of the three PID terms contributes to the action based on its individual gain values $k_\text{P}$, $k_\text{I}$, and $k_\text{D}$ for proportional, integral, and derivative terms, respectively.
Tuning these terms is the central task in PID controller design.
Hence, a huge number of literature exists on how to tune PID controllers for various application domains.

The most common prescriptive rules used in manual PID tuning are the two Ziegler-Nichols response methods~\citep{ziegler:42}.
While the first method is applied to plants with step responses, the second method targets unstable plants in a closed loop system, as considered in this work.

In the CPB problem, the two process variables $\theta$ and $\rho$ have to be regulated to their respective setpoints $\theta=0$ and $\rho=0$.
To achieve this goal using PID control theory, a controller layout consisting of two independent PID controllers can be utilized~\citep{wang:11}.
The error values $e_\theta$ and $e_\rho$ are determined separately before they are passed on to the respective PID terms.
The terms for each component of the controller are computed as follows:
\begin{equation}
\text{P}_s = k_{\text{P}_s} e_s,\ \text{I}_s = k_{\text{I}_s} \int_{t-T1}^{t}e_s dt,\ \text{D}_s = k_{\text{D}_s} \frac{de_s}{dt},
\end{equation}
with $s$ as a placeholder for $\theta$ and $\rho$.
Hence, the final output of the controller is determined by
\begin{equation}
a(t)=k_\theta(\text{P}_\theta+\text{I}_\theta+\text{D}_\theta)+k_\rho(\text{P}_\rho+\text{I}_\rho+\text{D}_\rho),
\end{equation}
where $k_\theta$ and $k_\rho$ are gains for balancing the contributions from $\theta$ and $\rho$, respectively.
Note that tuning this PID controller for CPB using the manual method from Ziegler and Nichols is a feasible yet laborious process.
It requires multiple simulation runs using a surrogate model of the CPB and subsequent investigations of the produced trajectories in order to converge to an adequate control performance.
	
	\subsection{Neural fitted Q iteration (NFQ)}
\label{section:nfq}

NFQ is an algorithm for efficient and effective training of an action-value function (Eq.~\eqref{eq:action_value_function_update_step}).
NFQ uses the ability of neural networks (NNs) to approximate non-linear functions to represent a value function~\citep{riedmiller:051}.
A huge advantage of using a global function representation, like with NNs, is that it can exploit generalization effects by assigning similar values to related areas. 
However, since changing the weights of NNs has a global effect on the value function output, many difficulties have been reported~\citep{boyan:95}.
In classical online RL, the value function is updated as soon as a new state-action experience has been made.
However, adopting the weights of the NN according to this recent sensation usually not only changes the value of this state-action pair and its related area, but also has an unpredictable influence on values of unrelated state-action examples.
With NFQ the malicious influence of a new update is constraint by offering previous knowledge explicitly.

In its original form, NFQ is a model-free RL approach.
This means that no explicit model of the environment has to be learned, which makes the method data-efficient for many applications since transition tuples from the plant are directly used to find well-performing control policies.
Generally, the method is also applicable for batch RL problems, where policies cannot be evaluated during the training and consequently cannot generate new transition samples.
However, using an environment model to evaluate and select adequate policies during training is very useful in order to yield adequate policy performance.

To adapt Q-learning for NNs, an error function can be derived from Eq.~\eqref{eq:action_value_function_update_step} to measure the difference between $q_{k}(\bm{s}_t,\bm{a}_t)$ and $q_{k+1}(\bm{s}_t,\bm{a}_t)$:
\begin{equation}
e_{k+1}=\left(q_{k+1}(\bm{s}_t,\bm{a}_t)-\left(\bm{r}_{t+1}+\gamma\max_{\bm{a}_{t+1}}q_{k}(\bm{s}_{t+1},\bm{a}_{t+1})\right)\right)^2.
\label{eq:nfq_learning}
\end{equation}
Note that minimizing this mean squared error by common gradient descent techniques for adjusting the weights after each new sample can cause unpredictable effects on other areas in the state-action space.
Therefore, \citeauthor{riedmiller:051} proposed to update the neural function offline on the entire set of previous transitions~\citep{riedmiller:051}.

The result of NFQ is an action-value function which has to be maximized over the set of available actions w.r.t. the current state.
Since this action-value function is represented by a multi-layer NN, it is generally very difficult to interpret how a certain action has been computed.
	
	\subsection{Particle swarm optimization policy (PSO-P)}
\label{section:psop}

PSO-P is a heuristic for solving RL problems by employing numerical online optimization of control action sequences~\citep{hein:16}. 
As an initial step, a system model is trained from observational data with standard methods. 
The problem of finding optimal control action sequences based on model predictions is solved with PSO, an established algorithm for non-convex optimization which does not require any gradient information~\citep{kennedy:95}. 
Specifically, PSO-P iterates over the following steps. 
(i) PSO is employed to search for an action sequence that maximizes the expected return when applied to the current system state by simulating its effects using the system model. 
(ii) The first action of the sequence with the highest expected return is applied to the real-world system. 
(iii) The system transitions to the subsequent state, and the optimization process is repeated based on the new state (go to step (i)).

The goal of PSO-P is to find an action sequence $\bm{x}=(\bm{a}_t,\bm{a}_{t+1},\ldots,$ $\bm{a}_{t+T-1})$ that maximizes the expected return $\mathcal{R}$ for state $\bm{s}_t$. 
Adopting the model-based value estimation from Eq.~\eqref{eq:model_based_value}, the expected return $\mathcal{R}_{\bm{x}}$ of action sequence $\bm{x}$ starting from state $\bm{s}_t$ is computed by the following value function:
\begin{align}
v_{\bm{x}}(\bm{s}_t) =& \sum_{k=0}^{T-1} \gamma^k r(\bm{s}_{t+k},\bm{a}_{t+k},\bm{s}_{t+k+1}),\\
\text{with }\bm{s}_{t+k+1}=&g(\bm{s}_{t+k},\bm{a}_{t+k}).
\label{equation:psop_return}
\end{align}

Solving the RL problem corresponds to finding the optimal action sequence $\hat{\bm{x}}$ by maximizing
\begin{equation}
\label{equation:psop_fitness}
\hat{\bm{x}}\in\argmax_{\bm{x}\in\mathcal{A}^T}v_{\bm{x}}(\bm{s}_t).
\end{equation}

	\subsection{Particle swarm optimization neural network (PSONN)}
\label{section:psonn}

In PSONN, the policy is represented by an NN whose weights are optimized by PSO. 
The inputs of the NN policy are the state features and the output is the respective action.
Using PSO to optimize the weights of NNs has been successfully applied in several publications~\citep{zhang:00,juang:04}.

Herein, the networks' topologies, as well as the neurons' transfer functions, are fixed and not part of the swarm optimization.
Parameter vector $\bm{x}$ defines the weights of the network graph edges which interconnect each neuron of one layer with all the neurons of the next layer.
Each particle of the swarm represents one solution within the parameter space.
Using model-based RL, the swarm searches for parameters yielding a policy with maximum average return (Eq.~\eqref{eq:return}).

The PSONN policy $\pi[\bm{x}](\bm{s})=\bm{a}$ receives a state $\bm{s}$ as input, transforms the state features' values according to the pre-defined activation functions and weights in $\bm{x}$, and passes the neurons' outputs further on to the next layer in the network.
The signals produced by the final layer are interpreted as action $\bm{a}$, which is subsequently applied to the system.
	
	\subsection{Fuzzy particle swarm reinforcement learning (FPSRL)}
\label{section:fpsrl}

FPSRL utilizes a model-based RL approach similar to PSONN, but instead of optimizing the weights of an NN policy, PSO is used to search for optimal fuzzy-rule-based policy parameters~\citep{hein:17c}.

Based on fuzzy set theory, \citeauthor{mamdani:75} introduced a so-called fuzzy controller, specified by a set of linguistic if-then rules whose membership functions can be activated independently and produce a combined output computed by a suitable defuzzification function~\citep{mamdani:75}.

In a $D$-inputs-single-output system with $C$ rules, a fuzzy rule $R^{(i)}$ can be expressed as follows:
\begin{equation}
R^{(i)}: \text{ IF }\bm{s}\text{ is } m^{(i)} \text{ THEN }o^{(i)}, \quad \text{with }i\in\{1,\dotsc,C\}, 
\end{equation}
where $\bm{s}\in \mathbb{R}^{D}$ denotes the input vector (the environment state in our setting), $m^{(i)}$ is the membership of a fuzzy set of the input vector in the premise part, and $o^{(i)}$ is a real number in the consequent part.

Herein, Gaussian membership functions~\citep{wang:92} have been applied:
\begin{equation}
m^{(i)}(\bm{s})=\text{m}[\bm{c}^{(i)},\bm{\sigma}^{(i)}](\bm{s})=\prod^{D}_{j=1}\exp\left\{-\frac{(c_{j}^{(i)}-s_j)^2}{2{\sigma_{j}^{(i)}}^2}\right\},
\label{gaussian_membership}
\end{equation}
where $m^{(i)}$ is the i-th parameterized Gaussian $\text{m}[\bm{c},\bm{\sigma}]$ with its center at $\bm{c}^{(i)}$ and width $\bm{\sigma}^{(i)}$.

The parameter vector $\bm{x}\in\mathcal X$, where $\mathcal X$ is the set of valid Gaussian fuzzy parameterizations, is represented by
\begin{align}
\label{eq:x_vector}
\begin{split}
\bm{x}=( & c_{1}^{(1)},c_{2}^{(1)},\dotsc,c_{D}^{(1)},\sigma_{1}^{(1)},\sigma_{2}^{(1)},\dotsc,\sigma_{D}^{(1)},o^{(1)},\\
& c_{1}^{(2)},c_{2}^{(2)},\dotsc,c_{D}^{(2)},\sigma_{1}^{(2)},\sigma_{2}^{(2)},\dotsc,\sigma_{D}^{(2)},o^{(2)},\dotsc,\\
& c_{1}^{(C)},c_{2}^{(C)},\dotsc,c_{D}^{(C)},\sigma_{1}^{(C)},\sigma_{2}^{(C)},\dotsc,\sigma_{D}^{(C)},o^{(C)},\alpha).
\end{split}
\end{align}

The output is determined using the following formula: 
\begin{equation}
\pi[\bm{x}](\bm{s})=\tanh\left(\alpha\cdot\frac{\sum_{i=1}^{C}m^{(i)}(\bm{s})\cdot o^{(i)}}{\sum_{i=1}^{C}m^{(i)}(\bm{s})}\right),
\label{eq:defuzzifier}
\end{equation}
where the hyperbolic tangent limits the output to between -1 and 1, and parameter $\alpha$ can be used to change the slope of the function.
The normalization applied in Eq.~\eqref{eq:defuzzifier} in combination with the Gaussian membership function ensures that even a relatively small number of membership functions can produce policy outputs for all possible inputs.
	
	\subsection{[Fuzzy] Genetic programming reinforcement learning ([F]GPRL)}
\label{section:fgprl}

Applying the aforementioned FPSRL to systems with many state features is prone to yield non-interpretable fuzzy systems since every rule contains all the state dimensions, including redundant or irrelevant ones, in its membership function by default.
By creating fuzzy rules using GP rather than tuning the fuzzy rule parameters via PSO, FGPRL eliminates the manual feature selection process~\citep{hein:18b}.
GP is able to automatically select the most informative features as well as the most compact fuzzy rule representation for a certain level of performance.
Moreover, it returns not just one solution to the problem but a whole Pareto front containing the best-performing solutions for many different levels of complexity.

GP has been utilized for creating system controllers since its introduction~\citep{koza:92}. 
Since then, the field of GP has grown significantly and has produced numerous results that can compete with human-produced solutions~\citep{koza:10}, including system controllers~\citep{keane:02,shimooka:98}, game playing~\citep{gearhart:03,wilson:18}, and robotics~\citep{downing:01,kamio:05}. 

\begin{figure}
	\centering
	\includegraphics[width=2.85in]{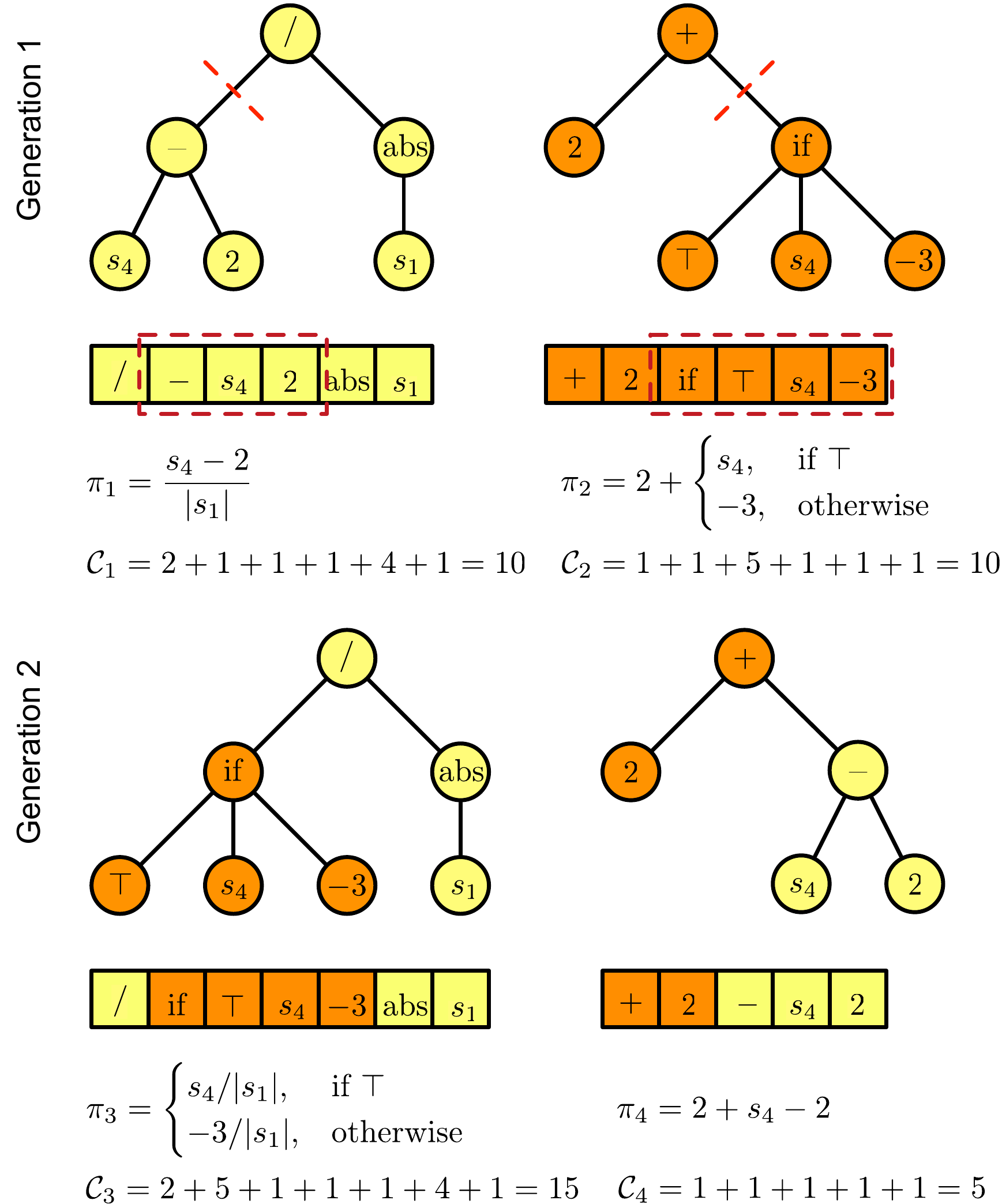}
	\caption[GP individuals as function trees and memory arrays]{GP individuals as function trees and memory arrays.
		Depicted are four exemplary GP individuals $\pi_i$ from two consecutive generations with their respective complexity measures $\mathcal{C}_i$.
		Crossover cutting points are marked in the tree diagrams of policies $\pi_1$ and $\pi_2$.}
	\label{figure:gp_trees}
\end{figure}

GP encodes computer programs as sets of genes and then modifies (evolves) them using a so-called genetic algorithm (GA) to drive the optimization of the population by applying selection and reproduction to the population.
The basis for both concepts is a fitness value which represents the quality of performing the predefined task for each individual.
Selection means that only the best portion of the current generation will survive each iteration and continue existing in the next generation.
Analogous to biological sexual breeding, two individuals are selected for reproduction based on their fitness, and two offspring individuals are created by crossing their chromosomes.
Technically, this is realized by selecting compatible cutting points in the function trees and subsequently interchanging the subtrees beneath these cuts.
The two resulting individuals are introduced to the population of the next generation (Fig.~\ref{figure:gp_trees}).
Herein, we applied tournament selection~\citep{blickle:95} for selecting the individuals to be crossed.

Since with FGPRL we are interested in using interpretable fuzzy controllers as RL policies, the genes include membership and defuzzification functions, as well as constant floating-point numbers and state variables. 
These fuzzy policies can be represented as function trees and stored efficiently in arrays~\citep{hein:18b}.

The interpretable policies which are generated by applying the GPRL approach are basic algebraic equations~\citep{hein:18c}.
Given that GPRL can find rather short (non-complex) equations, it is expected to reveal substantial knowledge about underlying coherencies between available state variables and well-performing control policies for a certain RL problem.

To rate the quality of each policy candidate, a fitness value has to be provided for the GP algorithm to advance.
For FGPRL and GPRL, the fitness of each individual is calculated by generating trajectories using the model-based return estimation from Eq.~\eqref{eq:return}.
	
	\section{Experiments}
	\label{section:experiments}
	
	\subsection{Cart-pole balancing (CPB) benchmark}
\label{section:benchmark}

The cart-pole experiments were conducted using the $CLS^2$ software\footnote{\url{http://ml.informatik.uni-freiburg.de/former/research/clsquare.html}} with default application parameters.
The objective of the CPB benchmark is to apply forces to a cart moving on a one-dimensional track to keep a pole hinged to the cart in an upright position. 
Here, the four Markov state variables are the pole angle $\theta$, the pole angular velocity $\dot\theta$, the cart position $\rho$, and the cart velocity $\dot\rho$. 
These variables describe the Markov state completely, i.e., no additional information about the system's past behavior is required. 
The task for the CPB controller is to find a sequence of force actions $a_t,a_{t+1},a_{t+2},\ldots$  that prevent the pole from falling over.

In the CPB task, the angle of the pole and the cart's position are restricted to intervals of $[-0.7,0.7]$ and $[-2.4,2.4]$ respectively. 
Once the cart has left the restricted area, the episode is considered a failure, i.e., velocities become zero, the cart's position and pole's angle become fixed, and the system remains in the failure state for the rest of the episode. 
The controller can apply force actions on the cart from $-10$~N to $+10$~N in time intervals of $0.025$~s.

The reward function (negative cost function) for CPB is given as follows:
\begin{equation}
    r(\bm{s}_{t+1})=
    \begin{cases}
        0.0, & \text{if }\abs{\theta_{t+1}}<0.25\text{ and } \abs{\rho_{t+1}}<0.5,\\
        -1.0, & \text{if }\abs{\theta_{t+1}}>0.7\text{ or } \abs{\rho_{t+1}}>2.4,\\
        -0.1, & \text{otherwise}.
    \end{cases}
    \label{eq:reward}
\end{equation}
Based on this reward function, the primary goal of the policy is to avoid reaching the failure state. 
The secondary goal is to drive the system to the goal state region where $r=0$ and balance it there for the rest of the episode.

In the experiments in the following section, discount factor $\gamma=0.97$ and time horizon $T=100$ have been used.
All of the methods utilized the same previously generated transition data batch of size $10\,000$.

\subsection{Neural network surrogate model}
\label{section:model}

The required model in Eq.~\eqref{eq:model} is an NN of the form 5-10-10-10-1 (from input to output) with rectifier activation functions on the three hidden layers.
The model computes the deltas of the state variables $\Delta \bm{s}_{t+1}$ from which the resulting state is calculated according to $\bm{s}_{t+1}=\bm{s}_{t}+\Delta \bm{s}_{t+1}$.
Since CPB yields discrete rewards, one-hot encoding, which transforms the categorical rewards (Eq.~\eqref{eq:reward}) to a more conveniently predictable problem representation, is used.
The rewards are mapped to the following binary vectors: $0.0\rightarrow[1,0,0]$,$-0.1\rightarrow[0,1,0]$, and $-1.0\rightarrow[0,0,1]$.
Note that such a representation can be interpreted as a probability vector for each observed reward with respect to one of the reward classes.
Consequently, the NN does not predict reward $r$ directly but it is rather optimized to predict the probabilities for each reward class by means of the one-hot vector.
	
	\subsection{Results and comparison}
\label{section:results}

LQR produced the following linear controller:
\begin{equation}
	\pi(\bm{s}_t) = -\bm{K}\bm{s}_t=38.8\theta+10.1\dot{\theta}+2.8\rho+3.9\dot{\rho}.
	\label{equation:lqr_solution}
\end{equation}
The penalty of this solution was $2.23$ on the model and $3.01$ on the real system, which represents a very good control performance for the CPB task.

\begin{figure}
	\begin{center}
		\includegraphics[width=3.4in]{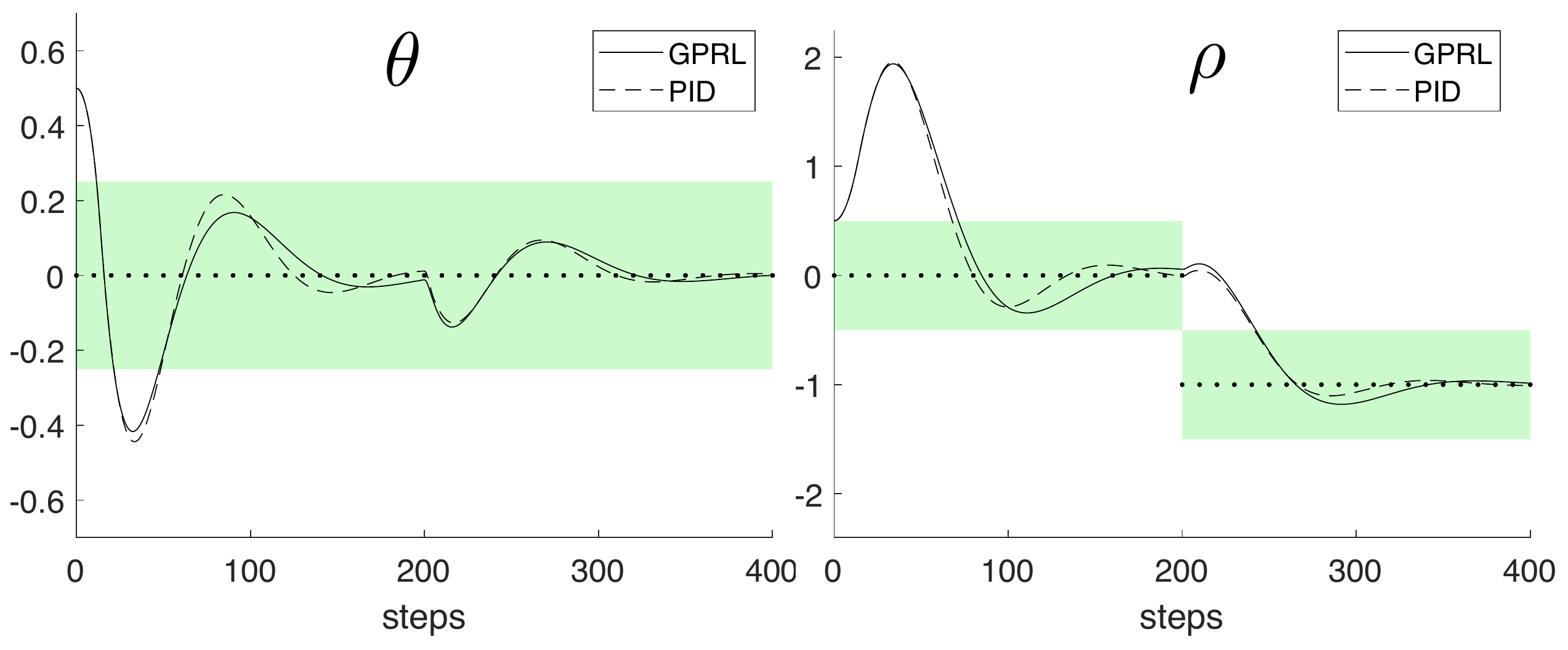}
		\caption{CPB trajectories for PID and GPRL. Both controllers successfully react to the change in setpoint and the respective goal area after 200 steps.}
		\label{figure:cpb_trajectoriesl}
	\end{center}
\end{figure}

The best controller found by the Ziegler-Nichols tuning is represented by:
\begin{align}
\begin{split}
a(t)= & 0.95\bigg(0.6k_\text{C1}\cdot e_\theta(t)
+1.2k_\text{C1}/p_\text{C1}\cdot \int_{t-40}^{t}e_\theta(t)dt\\
&+0.6k_\text{C1}p_\text{C1}/8\cdot \frac{de_\theta(t)}{dt}\bigg)\\
&+0.05\bigg(0.6k_\text{C2}\cdot e_\rho(t)
+1.2k_\text{C2}/p_\text{C2}\cdot \int_{t-28}^{t}e_\rho(t)dt\\
&+0.6k_\text{C2}p_\text{C2}/8\cdot \frac{de_\rho(t)}{dt}\bigg),
\end{split}
\end{align}
with critical values $k_\text{C1}=-14.3$ and $k_\text{C2}=-42.7$, and critical periods $p_\text{C1}=113$ and $p_\text{C2}=331$.
The penalty of this solution was $2.03$ on the model and $3.04$ on the real system.
Fig.~\ref{figure:cpb_trajectoriesl} depicts an example trajectory of this PID controller.
Note how it quickly drives the system into the goal regions and adjusts its action to the changed cart position setpoint after 200 steps.

\begin{figure}
	\begin{center}
		\includegraphics[width=3.4in]{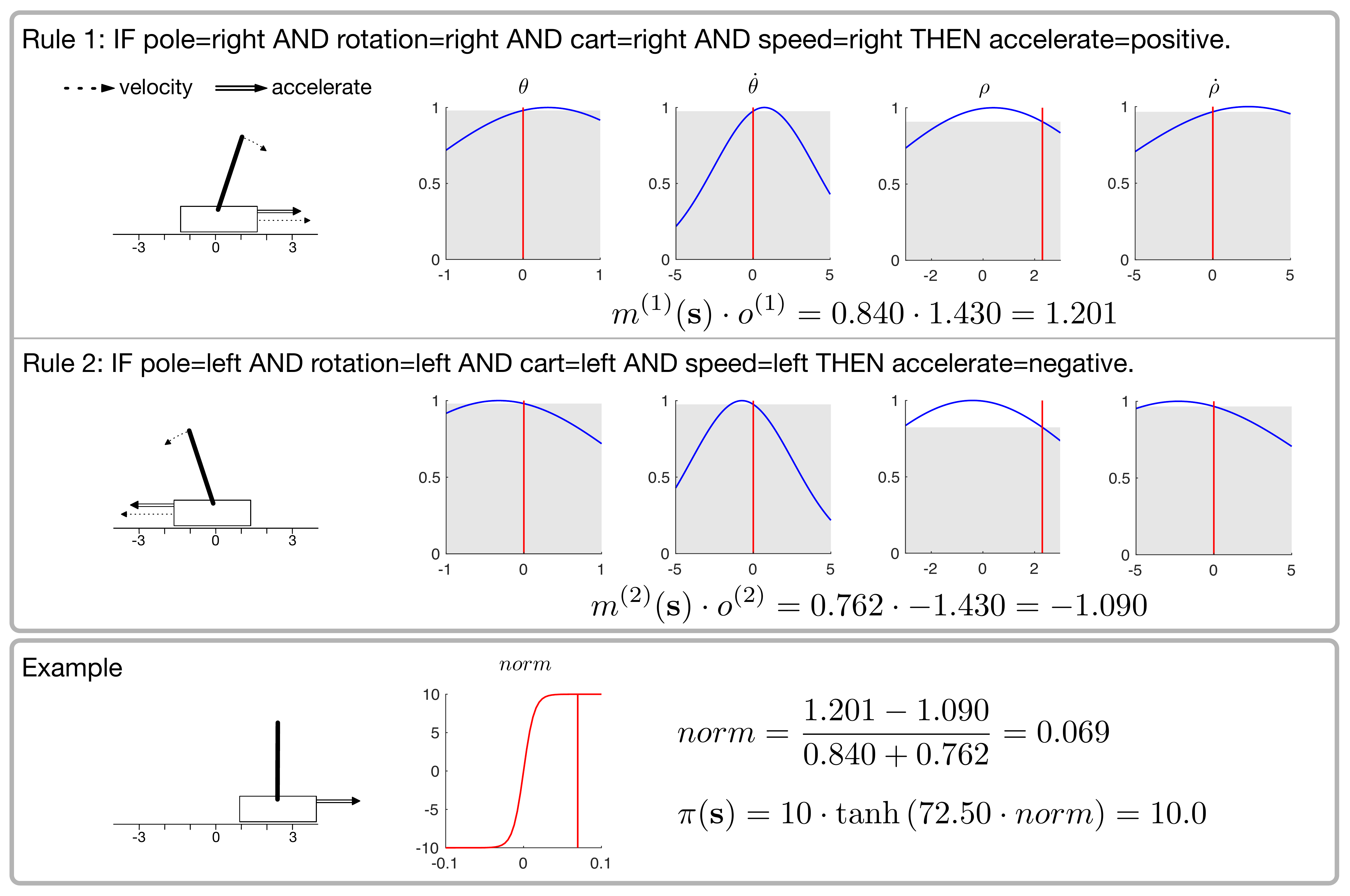}
		\caption[CPB rules by FPSRL]{CPB rules by FPSRL}
		\label{figure:rules_cpb}
	\end{center}
\end{figure}

Fig.~\ref{figure:rules_cpb} depicts a fuzzy rule policy generated by FPSRL.
The penalty of this solution was $2.02$ on the model and $2.72$ on the real system.
Only two interpretable fuzzy rules are sufficient to solve the CPB task.
However, allowing for higher complexity by tuning the parameters of 4 and 6 fuzzy rules increased the performance up to median penalties of $1.82$ on the model and $2.69$ on the real system dynamics.

\begin{figure}
	\begin{center}
		\includegraphics[width=3.4in]{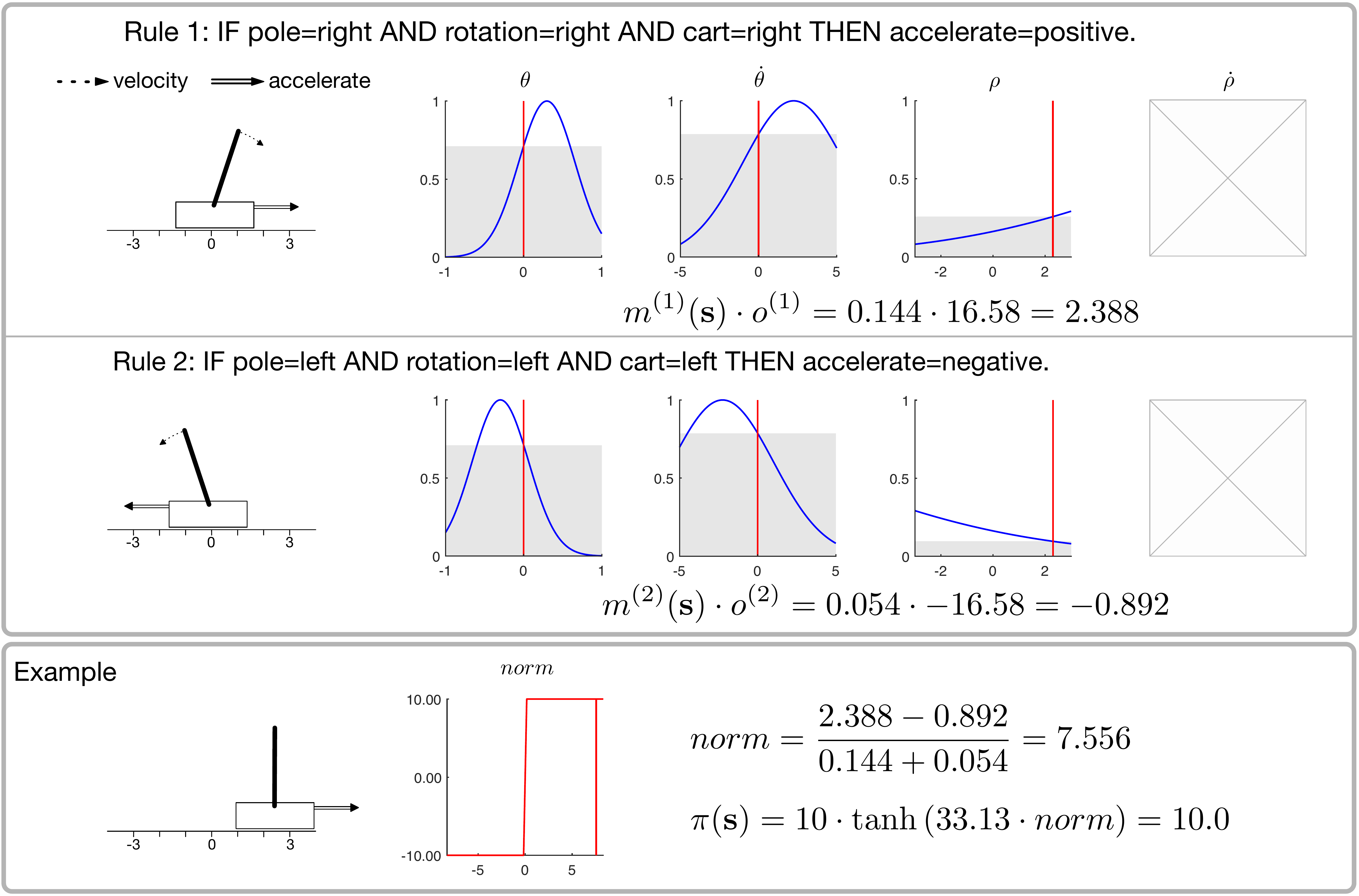}
		\caption[CPB rules by FGPRL]{CPB rules by FGPRL}
		\label{figure:rules_cpb_fgprl}
	\end{center}
\end{figure}

In contrast to FPSRL, FGPRL has the possibility to reduce the complexity by automatically selecting only the most relevant input features.
Fig.~\ref{figure:rules_cpb_fgprl} depicts a fuzzy rule policy generated by FGPRL, where only three out of four input features are used.
The penalty of this solution was $2.31$ on the model and $3.30$ on the real system, which means a small decline in performance compared to the more complex FPSRL solution.
Note that FGPRL produces a whole Pareto front (between the functions performance and complexity) of solutions for many levels of complexity incl. the solution with two rules and four input features.
The performance of the best policies irrespective of their complexity is shown in Table~\ref{table:compare}.

A solution of low complexity and high performance generated by GPRL is the following linear equation:
\begin{equation}
	\pi(\bm{s})=6.98\theta+2\dot{\theta}+\rho+0.94\dot{\rho}.
\end{equation}
The penalty of this solution was $2.00$ on the model and $2.73$ on the real system.
Note the similarities with the LQR solution from Eq.~\eqref{equation:lqr_solution}.
The ratios between $\theta$ and $\dot{\theta}$ are $3.8$ and $3.5$ for LQR and GPRL, respectively, while $\rho$ and $\dot{\rho}$ are weighted by significantly smaller gains.
Fig.~\ref{figure:cpb_trajectoriesl} compares GPRL trajectories for an example start state with that of the PID controller.
Note how the fully automatically generated solution from GPRL yields a very similar control performance compared to the manually tuned PID controller.
Similarly to FGPRL, GPRL generates a whole Pareto front of solutions.
The performance of the best policies is given in Table~\ref{table:compare}.

Since the policies of NFQ, PSO-P, and PSONN are non-interpretable, only their performance is compared with the other methods in Table~\ref{table:compare}.
PSO-P achieves outstanding performance values as long as its actions are applied to the same model on which the optimization is performed on.
Applying the actions to the real system dynamics  produces high penalty values.
NFQ in a pure model-free setting (\textit{last}) was not able to produce well-performing policies.
However, using a surrogate model for selecting the policy (\textit{selected}) yields adequate results.
The population-based RL method PSONN produced very good non-interpretable NN controllers for the CPB benchmark.
Parameters for all the compared methods evaluated on CPB are published in~\cite{hein:19}.

\begin{table*}
	\centering
	\resizebox{\textwidth}{!}{
		\begin{tabular}{clcrrrrrrrr}
			\hline
			& \multicolumn{1}{c}{LQR}  & \multicolumn{1}{c}{PID}  & \multicolumn{1}{c}{PSO-P} & \multicolumn{1}{c}{NFQ} & \multicolumn{1}{c}{NFQ} & \multicolumn{1}{c}{PSONN} & \multicolumn{1}{c}{FPSRL} & \multicolumn{1}{c}{FGPRL} & \multicolumn{1}{c}{GPRL}\\
			& & & & \multicolumn{1}{c}{(last)} & \multicolumn{1}{c}{(selected)} & & & &\\
			\hline
			Model  & $2.23$ & $2.0285$ & $1.2535\pm0.0014$  & $15.0\pm1.0$ & $2.71\pm0.04$ & $1.833\pm0.002$ & $1.819\pm0.005$ & $1.814\pm0.003$ & $1.88\pm0.02$\\
			System  & $3.01$ & $3.0376$ & $4.20\pm0.04$  & $15.5\pm0.9$ & $5.4\pm0.2$ & $2.831\pm0.003$ & $2.69\pm0.03$ & $2.798\pm0.006$ & $2.71\pm0.03$\\
			\hline    
		\end{tabular}
	}
	\caption{
		Penalty comparison of all evaluated methods.
		Depicted are the mean penalties computed from ten independent experiments.
		The number after $\pm$ represents the standard error.
		LQR and PID controllers are single manually constructed solutions.
	}
	\label{table:compare}
\end{table*}
	
	\subsection{Evaluation on a hardware cart-pole system}
\label{section:hardware}

To further evaluate the performance of population-based RL and to demonstrate the applicability to real dynamic systems, we applied GPRL on a hardware cart-pole system.
This cart-pole system (Fig.~\ref{figure:cebit_pendel}) is built from inexpensive components, like an ultrasonic sensor (Elegoo  HC-SR04) and a low-cost angle sensor (pwb encoders MEC 22).
The inexact information provided by these sensors yields noisy observations from the system.
Moreover, the velocities can only be estimated from differences of the positions.
Thus, the state of the hardware cart-pole is no longer a true Markovian state, which means we have to estimate the Markovian state from past observations.
The estimated state is given by $\tilde{\bm{s}}_t=(\theta_{t-9},\dot{\theta}_{t-9},\rho_{t-9},\dot{\rho}_{t-9},a_{t-9},\ldots,\theta_{t},\dot{\theta}_{t},\rho_{t},\dot{\rho}_{t},a_t)$, which yields 50 available input features for a policy.

Approximately $17\,000$ transition samples have been generated by a human player driving the cart by using a game pad.
Two NNs $\tilde{g}_{\theta}(\tilde{\bm{s}}_t)=\dot{\theta}_{t+1}$ and $\tilde{g}_{\rho}(\tilde{\bm{s}}_t)=\dot{\rho}_{t+1}$ are learned by supervised ML.
Angle and position are updated by computing $\theta_{t+1}=\theta_{t}+\dot{\theta}_{t+1}$ and $\rho_{t+1}=\rho_{t}+\dot{\rho}_{t+1}$, respectively.

A controller found by GPRL is represented by the following human-interpretable equation:
\begin{equation}
	\pi(\tilde{\bm{s}}_t) = 25\theta_t - \frac{\theta_{t-9}}{0.65+\dot{\theta}_{t-8}} + \rho_{t-7} + \dot{\rho}_{t-1}.
\end{equation}
It is clearly visible that the controller has a strong proportional term w.r.t. $\theta$ but also utilizes a past value of cart position $\rho$.

A video showing the data generation process and the learned controller in action is available at \url{https://www.youtube.com/watch?v=5ot3xUyFPq4}.

	\section{Conclusion}
\label{section:conclusion}

The experiments with both a CPB software simulation and a hardware demonstrator have shown that the population-based RL methods FPSRL, FGPRL, and GPRL are able to generate human-interpretable controllers from available transition samples.
In contrast to LQR and PID controllers which are only linear, symmetric around the setpoint, do not incorporate planning, and
cannot utilize direct knowledge of the process, RL is theoretically capable of solving any Markov decision process (MDP) which makes it suitable for a broader field of application.

In real-world industry use cases, data batches of transitions generated by default controllers are often available, whereas online RL is prohibited for safety reasons. 
Therefore, all of the investigated RL methods are offline approaches which learn directly from batch data or learn a surrogate model first using supervised learning and subsequently perform Monte Carlo rollouts to estimate the value function.
Since the space of solutions of GP consists of computer programs, it can search a wide variety of controller classes for domain specific use cases.
Examples include precisely defined signal plans for traffic control solutions or optimal scheduling policies for manufacturing industries.
	
	
	\bibliography{bibliography}

\begin{thebibliography}{33}
\providecommand{\natexlab}[1]{#1}
\providecommand{\url}[1]{\texttt{#1}}
\expandafter\ifx\csname urlstyle\endcsname\relax
  \providecommand{\doi}[1]{doi: #1}\else
  \providecommand{\doi}{doi: \begingroup \urlstyle{rm}\Url}\fi

\bibitem[Bellman(1962)]{bellman:62}
R.~E. Bellman.
\newblock \emph{Adaptive Control Processes: {A} Guided Tour}.
\newblock Princeton University Press, 1962.

\bibitem[Bertsekas(1995)]{bertsekas:95}
D.~P. Bertsekas.
\newblock \emph{Dynamic programming and optimal control}, volume~1.
\newblock Athena scientific Belmont, MA, 1995.

\bibitem[Bischoff et~al.(2013)Bischoff, Nguyen-Tuong, Koller, Markert, and
  Knoll]{bischoff:13}
Bastian Bischoff, Duy Nguyen-Tuong, Torsten Koller, Heiner Markert, and Alois
  Knoll.
\newblock Learning throttle valve control using policy search.
\newblock In \emph{Joint European Conference on Machine Learning and Knowledge
  Discovery in Databases}, pages 49--64. Springer, 2013.

\bibitem[Blickle and Thiele(1995)]{blickle:95}
T.~Blickle and L.~Thiele.
\newblock A mathematical analysis of tournament selection.
\newblock In \emph{ICGA}, pages 9--16, 1995.

\bibitem[Boyan and Moore(1995)]{boyan:95}
Justin~A. Boyan and Andrew~W. Moore.
\newblock Generalization in reinforcement learning: {S}afely approximating the
  value function.
\newblock In \emph{Advances in neural information processing systems}, pages
  369--376, 1995.

\bibitem[Chang et~al.(2007)Chang, Hu, Fu, and Marcus]{chang:07}
Hyeong~Soo Chang, Jiaqiao Hu, Michael~C Fu, and Steven~I Marcus.
\newblock Population-based evolutionary approaches.
\newblock In \emph{Simulation-Based Algorithms for Markov Decision Processes},
  chapter~3, pages 61--87. Springer, 2007.

\bibitem[Chin and Jafari(1998)]{chin:98}
Hubert~H Chin and Ayat~A Jafari.
\newblock Genetic algorithm methods for solving the best stationary policy of
  finite {M}arkov decision processes.
\newblock In \emph{System Theory, 1998. Proceedings of the Thirtieth
  Southeastern Symposium on}, pages 538--543. IEEE, 1998.

\bibitem[Desborough and Miller(2002)]{desborough:02}
Lane Desborough and Randy Miller.
\newblock Increasing customer value of industrial control performance
  monitoring - {H}oneywell's experience.
\newblock In \emph{AIChE symposium series}, number 326, pages 169--189. New
  York; American Institute of Chemical Engineers; 1998, 2002.

\bibitem[Doshi-Velez and Kim(2017)]{doshi:17}
Finale Doshi-Velez and Been Kim.
\newblock Towards a rigorous science of interpretable machine learning.
\newblock \emph{arXiv preprint arXiv:1702.08608}, 2017.

\bibitem[Downing(2001)]{downing:01}
Keith~L. Downing.
\newblock Adaptive genetic programs via reinforcement learning.
\newblock In \emph{Proceedings of the 3rd Annual Conference on Genetic and
  Evolutionary Computation}, GECCO'01, pages 19--26, San Francisco, CA, USA,
  2001. Morgan Kaufmann Publishers Inc.

\bibitem[Ernst et~al.(2005)Ernst, Geurts, and Wehenkel]{ernst:05}
Damien Ernst, Pierre Geurts, and Louis Wehenkel.
\newblock Tree-based batch mode reinforcement learning.
\newblock \emph{Journal of Machine Learning Research}, 6:\penalty0 503--556,
  2005.

\bibitem[EU(2016)]{eu:16}
EU.
\newblock {Regulation (EU) 2016/679 of the European Parliament and of the
  Council of 27 April 2016 on the protection of natural persons with regard to
  the processing of personal data and on the free movement of such data, and
  repealing Directive 95/46/EC (General Data Protection Regulation)}.
\newblock \emph{OJ}, L 119:\penalty0 1--88, 5 2016.

\bibitem[Gearhart(2003)]{gearhart:03}
Chris Gearhart.
\newblock Genetic programming as policy search in {M}arkov decision processes.
\newblock In J.~R. Koza, editor, \emph{Genetic Algorithms and Genetic
  Programming at Stanford}, pages 61--67. Stanford Bookstore, Stanford, USA,
  2003.

\bibitem[Gomez et~al.(2006)Gomez, Schmidhuber, and Miikkulainen]{gomez:06}
Faustino Gomez, J{\"u}rgen Schmidhuber, and Risto Miikkulainen.
\newblock Efficient non-linear control through neuroevolution.
\newblock In \emph{European Conference on Machine Learning}, pages 654--662.
  Springer, 2006.

\bibitem[Hein et~al.(2016)Hein, Hentschel, Runkler, and Udluft]{hein:16}
D.~Hein, A.~Hentschel, T.~A. Runkler, and Steffen Udluft.
\newblock Reinforcement learning with particle swarm optimization policy
  ({PSO-P}) in continuous state and action spaces.
\newblock \emph{International Journal of Swarm Intelligence Research (IJSIR)},
  7\penalty0 (3):\penalty0 23--42, 2016.

\bibitem[Hein et~al.(2017)Hein, Hentschel, Runkler, and Udluft]{hein:17c}
D.~Hein, A.~Hentschel, T.~A. Runkler, and S.~Udluft.
\newblock Particle swarm optimization for generating interpretable fuzzy
  reinforcement learning policies.
\newblock \emph{Engineering Applications of Artificial Intelligence},
  65:\penalty0 87--98, 2017.

\bibitem[Hein(2019)]{hein:19}
Daniel Hein.
\newblock \emph{Interpretable Reinforcement Learning Policies by Evolutionary
  Computation}.
\newblock Dissertation, Technical University Munich, Munich, 2019.

\bibitem[Hein et~al.(2018{\natexlab{a}})Hein, Udluft, and Runkler]{hein:18b}
Daniel Hein, Steffen Udluft, and Thomas~A. Runkler.
\newblock Generating interpretable fuzzy controllers using particle swarm
  optimization and genetic programming.
\newblock In \emph{Proceedings of the Genetic and Evolutionary Computation
  Conference Companion}, GECCO '18, pages 1268--1275, New York, NY, USA,
  2018{\natexlab{a}}. ACM.

\bibitem[Hein et~al.(2018{\natexlab{b}})Hein, Udluft, and Runkler]{hein:18c}
Daniel Hein, Steffen Udluft, and Thomas~A. Runkler.
\newblock Interpretable policies for reinforcement learning by genetic
  programming.
\newblock \emph{Engineering Applications of Artificial Intelligence},
  76:\penalty0 158--169, 2018{\natexlab{b}}.

\bibitem[Juang(2004)]{juang:04}
Chia-Feng Juang.
\newblock A hybrid of genetic algorithm and particle swarm optimization for
  recurrent network design.
\newblock \emph{IEEE Transactions on Systems, Man, and Cybernetics, Part B
  (Cybernetics)}, 34\penalty0 (2):\penalty0 997--1006, 2004.

\bibitem[Kamio and Iba(2005)]{kamio:05}
S.~Kamio and H.~Iba.
\newblock Adaptation technique for integrating genetic programming and
  reinforcement learning for real robots.
\newblock \emph{Trans. Evol. Comp}, 9\penalty0 (3):\penalty0 318--333, June
  2005.

\bibitem[Keane et~al.(2002)Keane, Koza, and Streeter]{keane:02}
Martin~A. Keane, J.~R. Koza, and Matthew~J. Streeter.
\newblock Automatic synthesis using genetic programming of an improved
  general-purpose controller for industrially representative plants.
\newblock In \emph{Proceedings of the 2002 NASA/DoD Conference on Evolvable
  Hardware (EH'02)}, EH '02, pages 113--123, Washington, DC, USA, 2002. IEEE.

\bibitem[Kennedy and Eberhart(1995)]{kennedy:95}
J.~Kennedy and R.~C. Eberhart.
\newblock Particle swarm optimization.
\newblock \emph{Proceedings of the IEEE International Joint Conference on
  Neural Networks}, pages 1942--1948, 1995.

\bibitem[Koza(1992)]{koza:92}
J.~R. Koza.
\newblock \emph{Genetic Programming: {O}n the Programming of Computers by Means
  of Natural Selection}.
\newblock MIT Press, Cambridge, MA, USA, 1992.

\bibitem[Koza(2010)]{koza:10}
J.~R. Koza.
\newblock Human-competitive results produced by genetic programming.
\newblock \emph{Genetic Programming and Evolvable Machines}, 11\penalty0
  (3):\penalty0 251--284, 2010.

\bibitem[Mamdani and Assilian(1975)]{mamdani:75}
E.~H. Mamdani and S.~Assilian.
\newblock An experiment in linguistic synthesis with a fuzzy logic controller.
\newblock \emph{Int. Journal of Man-Machine Studies}, 7\penalty0 (1):\penalty0
  1--13, 1975.

\bibitem[Riedmiller(2005)]{riedmiller:051}
Martin Riedmiller.
\newblock Neural fitted {Q} iteration - {F}irst experiences with a data
  efficient neural reinforcement learning method.
\newblock In \emph{Machine Learning: ECML 2005}, volume 3720, pages 317--328.
  Springer, 2005.

\bibitem[Shimooka and Fujimoto(1999)]{shimooka:98}
Hiroaki Shimooka and Yoshiji Fujimoto.
\newblock Generating equations with genetic programming for control of a
  movable inverted pendulum.
\newblock In \emph{Selected Papers from the Second Asia-Pacific Conference on
  Simulated Evolution and Learning on Simulated Evolution and Learning},
  SEAL'98, pages 179--186, London, UK, 1999. Springer-Verlag.

\bibitem[Wang(2011)]{wang:11}
Jia-Jun Wang.
\newblock {Simulation studies of inverted pendulum based on PID controllers}.
\newblock \emph{Simulation Modelling Practice and Theory}, 19\penalty0
  (1):\penalty0 440--449, 2011.

\bibitem[Wang and Mendel(1992)]{wang:92}
L.-X. Wang and J.~M. Mendel.
\newblock Fuzzy basis functions, universal approximation, and orthogonal
  least-squares learning.
\newblock \emph{IEEE Transactions on Neural Networks}, 3\penalty0 (5):\penalty0
  807--814, 1992.

\bibitem[Wilson et~al.(2018)Wilson, Cussat-Blanc, Luga, and Miller]{wilson:18}
Dennis~G Wilson, Sylvain Cussat-Blanc, Herv{\'e} Luga, and Julian~F Miller.
\newblock Evolving simple programs for playing {A}tari games.
\newblock In \emph{Proceedings of the Genetic and Evolutionary Computation
  Conference}, GECCO '18, pages 229--236, New York, NY, USA, 2018. ACM.

\bibitem[Zhang et~al.(2000)Zhang, Shao, and Li]{zhang:00}
Chunkai Zhang, Huihe Shao, and Yu~Li.
\newblock Particle swarm optimisation for evolving artificial neural network.
\newblock In \emph{Systems, Man, and Cybernetics, 2000 IEEE International
  Conference on}, volume~4, pages 2487--2490. IEEE, 2000.

\bibitem[Ziegler and Nichols(1942)]{ziegler:42}
John~G Ziegler and Nathaniel~B Nichols.
\newblock Optimum settings for automatic controllers.
\newblock \emph{Transactions of the A.S.M.E.}, 64\penalty0 (11), 1942.

\end{thebibliography}
	
\end{document}